%% file: neurips_2025.tex
\pdfoutput=1
\documentclass{article}

\usepackage[preprint]{neurips_2025}

\usepackage[utf8]{inputenc}     
\usepackage[T1]{fontenc}        
\usepackage{hyperref}           
\usepackage{url}                
\usepackage{booktabs}           
\usepackage{amsfonts}           
\usepackage{nicefrac}           
\usepackage{microtype}          
\usepackage[table,xcdraw,dvipsnames]{xcolor} 

\usepackage{amsmath}
\usepackage{colortbl}
\usepackage{bookmark}
\usepackage{graphicx}
\usepackage{caption}
\usepackage{tabularx}
\usepackage{xargs}
\usepackage{listings}
\usepackage{enumitem}
\usepackage{multirow}
\usepackage[normalem]{ulem}
\useunder{\uline}{\ul}{}

\usepackage[capitalize]{cleveref}
    \crefname{figure}{Figure}{Figures}
    \Crefname{figure}{Figure}{Figures}
    \crefname{section}{Section}{Sections}
    \Crefname{section}{Section}{Sections}
    \crefname{table}{Table}{Tables}
    \Crefname{table}{Table}{Tables}

\title{Interpretable and Reliable Detection of AI-Generated Images via Grounded Reasoning in MLLMs}

\author{%
  Yikun Ji \\
  Shanghai Jiao Tong University \\
  \And
  Hong Yan \\
  Ant Group \\
  \And
  Jun Lan \\
  Ant Group \\
  \And
  Huijia Zhu \\
  Ant Group \\
  \And
  Weiqiang Wang \\
  Ant Group \\
  \And
  Qi Fan \\
  Shanghai Jiao Tong University \\
  \And
  Liqing Zhang \\
  Shanghai Jiao Tong University \\
  \And
  Jianfu Zhang \\
  Shanghai Jiao Tong University \\
  \texttt{\{da-kun,c.sis\}@sjtu.edu.cn}
}

\begin{document}

\maketitle

\begin{abstract}
\input{sections/00-abstract}
\end{abstract}

\input{sections/01-intro}

\input{sections/02-related}
\input{sections/03-dataset}
\input{sections/04-method}
\input{sections/05-experiments}

\input{sections/07-conclusion}

\clearpage
\appendix
\input{sections/suppl}

\clearpage
\bibliographystyle{plain}
\bibliography{neurips_2025}

\end{document}

%% file: sections/00-abstract.tex
The rapid advancement of image generation technologies intensifies the demand for interpretable and robust detection methods. Although existing approaches often attain high accuracy, they typically operate as black boxes without providing human-understandable justifications. Multi-modal Large Language Models (MLLMs), while not originally intended for forgery detection, exhibit strong analytical and reasoning capabilities. When properly fine-tuned, they can effectively identify AI-generated images and offer meaningful explanations. However, existing MLLMs still struggle with hallucination and often fail to align their visual interpretations with actual image content and human reasoning. To bridge this gap, we construct a dataset of AI-generated images annotated with bounding boxes and descriptive captions that highlight synthesis artifacts, establishing a foundation for human-aligned visual-textual grounded reasoning. We then finetune MLLMs through a multi-stage optimization strategy that progressively balances the objectives of accurate detection, visual localization, and coherent textual explanation. The resulting model achieves superior performance in both detecting AI-generated images and localizing visual flaws, significantly outperforming baseline methods.

%% file: sections/01-intro.tex
\section{Introduction}
\label{sec:intro}

The past decade has witnessed rapid progress in text-to-image generation, evolving from Generative Adversarial Networks to Diffusion Models, which are now capable of producing images nearly indistinguishable from real photographs~\cite{GAN,DiT}. These advances have led to an explosion of highly realistic AI-generated content, raising pressing concerns about misinformation, authenticity, and trust in digital media. Most existing detection methods cast this task as a binary classification problem, leveraging convolutional neural networks or vision transformers~\cite{CNNSpot, NPR, ComFor}. However, binary authenticity labels offer limited insight into \textit{why} an image is classified as AI-generated. In real-world applications, especially those involving legal, journalistic, or ethical implications, explainable detection is essential. An effective detection system should not only identify whether an image is AI-generated but also pinpoint the specific visual cues or logical inconsistencies that betray its synthetic origin. Such explainability promotes user trust, supports verification workflows, and enables more informed decision-making.

The rise of Multi-modal Large Language Models (MLLMs) has enabled cross-modal inference, allowing models to generate human-readable reasoning about AI-generated images~\cite{R1-V, LMM-R1, VLM-R1}. Recent efforts~\cite{FakeBench,ABench} have advanced interpretable textual explanations using MLLMs.
However, without visual grounding, it remains uncertain whether these rationales accurately reflect image content or derive from hallucinations.
To enhance alignment and reduce hallucinations, fine-grained visual-textual supervision, such as region-level bounding boxes and detailed captions, is crucial. Yet, the scarcity of such datasets presents a key obstacle to developing trustworthy and interpretable MLLM-based image detection.

In this paper, we introduce the FakeXplained dataset, comprising high-quality AI-generated images, along with an MLLM fine-tuning pipeline that achieves state-of-the-art detection accuracy and grounding capability. Our approach provides comprehensive rationales for fake image detection that demonstrate performance comparable to human annotators, inaugurating a new paradigm in human-interpretable AI-generated image detection. Our major contributions are threefold:

\begin{itemize}[leftmargin=*]

    \item \textbf{FakeXplained dataset:} We curate a human-annotated dataset of 8,772 AI-generated images sourced from a diverse range of state-of-the-art generative models, annotating them with bounding boxes highlighting visual anomalies or illogical details, along with concise captions explaining the nature of each flaw. To the best of our knowledge, \textbf{FakeXplained} is the \textit{first} dataset to provide localized, textual explanations specifically for AI-generated image detection.

    \item \textbf{Two-stage MLLM fine-tuning:} We fine-tune \textit{Qwen-2.5-VL (32B)} on the FakeXplained dataset to build an end-to-end system capable of both detecting and explaining AI-generated images. Given an input image, the model produces (i) a binary authenticity decision (real or AI-generated), and (ii) for fake images, a set of predicted bounding boxes paired with natural language justifications for each identified region. After fine-tuning, the model achieves state-of-the-art performance across both classification and localization metrics.

    \item \textbf{Grounded reasoning:} Beyond quantitative improvements, we demonstrate that fine-tuning on \textbf{FakeXplained} enables MLLMs to perform fine-grained visual reasoning and clearly articulate their observations. The model is capable of answering not only \emph{``is it fake?''} but also \emph{``where and why does this image look fake?''}, producing interpretable, grounded explanations aligned with human judgment.

\end{itemize}

\begin{figure}[t]
    \centering
    \includegraphics[width=1\linewidth]{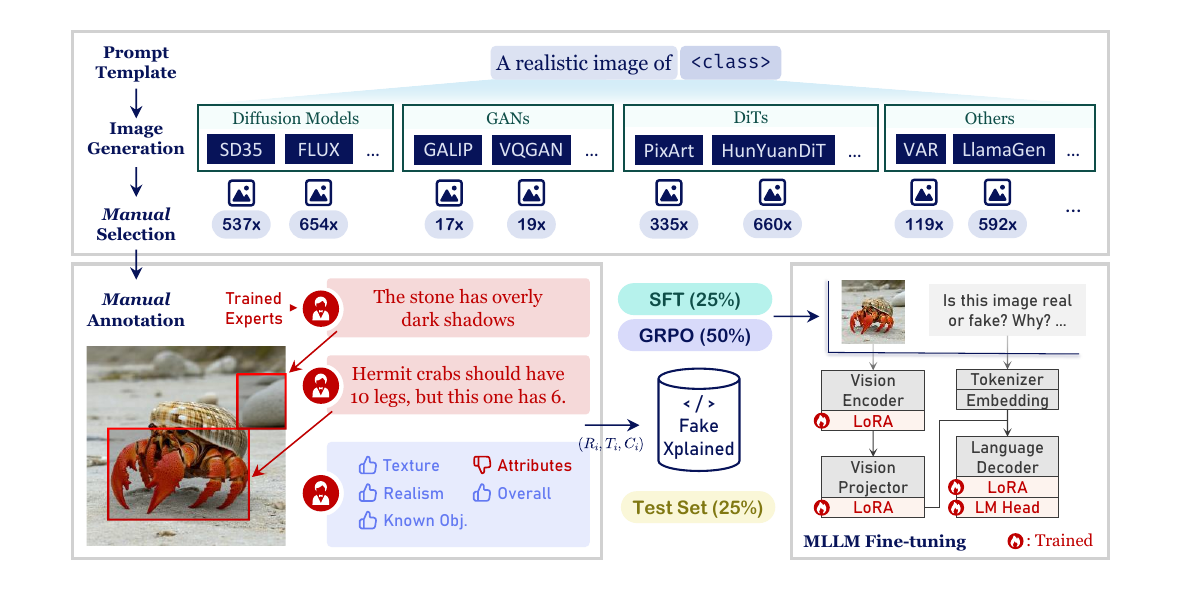}
    \caption{An overview of our method.}
    \label{fig:teaser}
\end{figure}

%% file: sections/02-related.tex
\section{Related works}
\label{sec:related}

\paragraph{Detection of AI-generated and manipulated images.}

Detecting AI-generated images has gained prominence with the improving fidelity of synthetic images from GANs~\cite{GAN,VQGAN}, autoregressive transformers~\cite{VQVAE}, diffusion-based models~\cite{SD21,SD3,DDIM,DDPM,VQDM,Imagen} and DiTs~\cite{DiT,PixArtAlpha}. Early research focused on artifact detection in spatial and frequency domains, targeting inconsistencies like upsampling artifacts~\cite{ArtifactUseFreq}, texture synthesis mismatches~\cite{ArtifactTextureDefake}, or limited high-frequency decay~\cite{ArtifactSpectrumDefake} to expose fakes, even imperceptible ones.

Beyond artifact-based approaches, deep learning methods such as ResNets and Vision Transformers trained on real and synthetic data~\cite{CNNSpot, EfficientNet, ComFor, AntifakePrompt} leverage strong feature extraction to learn discriminative patterns. However, generalization to unseen models remains challenging~{DefakeByReals}, and many approaches are vulnerable to domain-specific biases such as JPEG compression artifacts~\cite{FakeOrJPEG}. As generation techniques evolve, artifact-based cues alone become increasingly unreliable.
A complementary research direction focuses on model explainability, as most detectors offer only binary classification without indicating how or where synthetic cues are found. Recent efforts towards fine-grained or localized detection include measuring reconstruction errors~\cite{AEROBLADE}, using multi-branch systems for multi-level labels~\cite{DefakeByReals}, computing local intrinsic dimensionalities~\cite{IntrinsicDimensionalities}, or integrating text-image contrastive learning~\cite{XplainedAxiomaticAttr,XplainedCNN,XplainedGradCAM}. 
Benchmarks such as FakeBench~\cite{FakeBench} and A-Bench~\cite{ABench} aim to promote human-understandable textual explanations through a visual question answering (VQA)-style interface. However, without corresponding visual grounding, these explanations remain difficult to verify or trust. It is often unclear whether the textual rationale refers to any actual region in the image, and users are unable to determine whether the explanation is factually correct. This raises concerns about the factual alignment, interpretability, and credibility of such explanations.

\paragraph{Training \& fine-tuning reasoning-capable MLLMs}
Enhancing the reasoning capabilities of MLLMs is crucial for tasks requiring nuanced understanding. Initial strategies involved converting images into formalized textual representations to enable structured, language-driven reasoning~\cite{R1-OneVision}. Subsequent research has focused on instilling deeper cognitive abilities, including self-verification, self-correction, developing ``slow-thinking'' capabilities~\cite{VL-Rethinker}, and managing reasoning depth to address phenomena like ``overthinking''~\cite{Fast-Slow-Thinking}. Efforts also explore capturing human-like insight, or the ``aha moment,'' even in compact models~\cite{VisualThinker-R1-Zero}, and constructing high-quality multi-modal Chain-of-Thought (CoT) datasets~\cite{Vision-R1} to guide reasoning processes.

Reinforcement Learning (RL) has become pivotal in these advancements, with many sophisticated reasoning developments relying on RL methodologies. Fine-tuning methods such as Generative Reward Policy Optimization (GRPO)~\cite{DeepSeekMath}, introduced for MLLMs in work such as R1-V~\cite{R1-V}, have spurred significant interest in RL-based multi-modal reasoning. RL, particularly when combined with structured reward functions—for instance, using Intersection over Union (IoU) for tasks involving image grounding~\cite{VLM-R1}—markedly improves multi-modal alignment, visual reasoning, and human-interpretable decision-making. This has enabled progress across diverse vision-language applications, from generating interpretable analyses of medical images~\cite{MedVLM-R1} and enhancing text-to-image generation quality~\cite{T2I-R1}, to endowing smaller MLLMs with robust reasoning capacities through innovative RL frameworks~\cite{LMM-R1}. These approaches demonstrate RL's capability of advancing model performance in complex vision-language tasks.

%% file: sections/03-dataset.tex
\section{FakeXplained dataset}
\label{sec:dataset}

To address the hallucination problem and improve alignment between visual perception and human reasoning, we aim to train MLLMs not only to detect AI-generated images but also to articulate why they are fake in a reliable and human-understandable manner. This necessitates a dataset that supports both visual grounding and textual reasoning.

We introduce the \textbf{FakeXplained} dataset, designed for interpretable and trustworthy detection of AI-generated images. It consists of high-quality synthetic images paired with fine-grained, human-aligned annotations that indicate the underlying flaws and artifacts responsible for their detection as fake.

\subsection{Image generation}

To ensure data diversity, we generated images across 1,000 ImageNet categories using 28 different text-to-image generation models, as shown in \cref{tab:result-model-wise-acc-iou}. 
All generated images underwent manual quality screening.
Each image was independently rated by three volunteers, and the top-rated 8,772 AI-generated images were selected for subsequent annotation.

\subsection{Image annotation}
\label{subsec:image-annotation}
To support interpretable reasoning and help models understand what constitutes an AI-generated image, we provide detailed annotations for synthetic images. Real images are not annotated because they lack synthesis flaws.

\paragraph{Fake regions and explanations.}
Precise regional annotations and corresponding textual descriptions are essential for visual grounding and interpretability.
We recruited 23 trained annotators to label the high-quality AI-generated images selected from the previous stage. 
Their primary task was to identify and describe all regions within each image that exhibited signs of being fake.
Prior to annotation, all participants underwent standardized training focused on identifying visual cues of AI-generated content.
The training emphasized the identification of \textit{fake regions}, which are defined as areas within an image that either violate common sense or exhibit noticeable AI-generated artifacts.
Examples of common sense violations include anomalies such as ``a flamingo with three legs'' or ``bird feathers with a metallic appearance''.
Common AIGC artifacts include ``repetitive patterns on a blanket'' or ``blurred or illegible text''.

Annotators were also introduced to a structured annotation rubric to ensure consistency and alignment with the dataset’s objectives.
Each annotation consists of one or more fake regions, where each region is represented by a tuple $(R_i, T_i)$, where $R_i$ denotes a rectangular bounding box encapsulating the region, and $T_i$ provides a textual description of the identified anomaly or artifact.
On average, an annotated image in the dataset contains 5.42 such $(R_i, T_i)$ pairs,  which serve as the foundation for grounding and reasoning in downstream model training.

\paragraph{Image-level tagging.}
In addition to region-level annotation, annotators were asked to tag images based on broader perceptual attributes.
These attributes include texture quality, overall realism, correctness of attributes (\textit{e.g.}, whether depicted objects adhere to expected characteristics), recognizability of objects, and the presence of other significant defects not explicitly listed (\textit{e.g.}, the occurrence of multiple sub-images within a single image). 
These tags $C_i$ are mutually independent, allowing annotators to assign zero or multiple tags to each image as appropriate. 
This tagging framework allows the dataset to capture holistic image quality assessments, particularly in cases where visually realistic AI-generated images may lack distinct localized flaws.

\paragraph{Quality control.}
To ensure the reliability of the annotations, we implemented a quality control protocol involving both manual inspection and algorithmic validation. A subset of annotations was compared against a reference set of fake region annotations curated by the research team. 
Given the inherently subjective nature of visual interpretation, we adopted a tolerant validation criterion to accommodate diverse perspectives among annotators. Specifically, a minimum Intersection over Union (IoU) threshold of 20.0\% was applied for bounding box overlap, and an accuracy threshold of 33.3\% was used for image-level taggings. These metrics were assessed on a validation set comprising 5\% of the annotated images.

The IoU metric is used to assess the spatial agreement between annotated and reference bounding boxes.
Let $R_v$ represent the rectangular bounding box annotated by a volunteer, and $R_r$ represent the corresponding reference bounding box from the reference set. The Intersection over Union (IoU) is computed as:
$$
\text{IoU}(R_v, R_r) = \frac{|R_v \cap R_r|}{|R_v \cup R_r|},
$$
where $|R_v \cap R_r|$ denotes the area of the intersection between $R_v$ and $R_r$, and $|R_v \cup R_r|$ denotes the area of their union. The IoU value ranges from 0 to 1, with higher values indicating stronger alignment.
This quality control procedure ensures a baseline level of annotation fidelity while preserving the diversity of human interpretations. The resulting dataset, enriched with both region-level annotations $(R, T)$ and image-level tags $C$, offers a robust foundation for analyzing the semantic inconsistencies and perceptual flaws of AI-generated images.

%% file: sections/04-method.tex
\section{Methodologies}
\label{sec:method}

\subsection{Overview of the fine-tuning strategy}

Our goal is to fully leverage the constructed dataset to train MLLMs to not only determine whether an image is AI-generated, but also to localize the relevant regions and explain the rationale behind its prediction. The training pipeline follows a progressive optimization paradigm~\cite{DeepSeekMath}, beginning with supervised fine-tuning (SFT)~\cite{SFT} to establish formatting compliance and enable basic reasoning abilities, followed by Reinforcement Learning from Human Feedback (RLHF) implemented via progressive Group Relative Policy Optimization (GRPO).

Before training, each image's annotations are reformatted into an end-to-end dialogue between a user and an assistant, using a prompt structure designed for bounding-box-aware fine-tuning.
Inspired by recent studies in reasoning-augmented language models~\cite{openai-o1, deepseek-r1}, we adopt a structured output format that separates intermediate reasoning from final predictions.
Region-level annotations $(R_i, T_i)$ are enclosed within \verb|<think>| markers, image-level tags $C_i$ within \verb|<tag>| markers, and the final verdict is wrapped in \verb|<verdict>| markers.

\subsection{Cold start with supervised fine-tuning}
The cold start phase of fine-tuning uses SFT to establish a stable foundation before proceeding to RL.
During this phase, all linear layers of the vision encoder, projector, and language model components in the MLLM are fine-tuned based on the supervision signals from the data.
This initial fine-tuning is crucial for stabilizing the model prior to full-scale reinforcement learning training, preventing instabilities that might arise from pure RL-based updates~\cite{deepseek-r1}.

The SFT process focuses on teaching the model to produce coherent reasoning patterns with clear structure. The training emphasizes the consistent use of the designated marker format with \verb|<think>|, \verb|<tag>|, and \verb|<verdict>| fields, ensuring clarity and robustness in the model's reasoning outputs. This structured Chain-of-Thought format reduces errors and improves explainability, providing a solid foundation for subsequent GRPO stages that will refine the model's performance on specific metrics.

\subsection{Design of reward functions}
Reward design is a critical component of RLHF, guiding the MLLMs to learn not only how to detect fake images, but also how to localize relevant regions and provide coherent reasoning. Let $o$ denote the textual output of the MLLM. We define three core reward functions for this purpose.

\paragraph{Classification accuracy (\textit{Label}).}
To ensure the model produces the correct verdict, we extract the classification decision from within the \verb|verdict| marker and compare it with the ground-truth label:
$$\mathcal{R}_C(o) = \begin{cases} 1, \quad & \text{if } V(o) = y, \\ -1, \quad & \text{o.w.} \end{cases}$$
where $V(o)$ is a regex match for the verdict, and $y$ is the ground truth label of whether the image is real or generated.

\paragraph{Grounding accuracy (\textit{Relaxed IoU}).}
To reward alignment between model-predicted and human-annotated regions, we use a relaxed version of the Intersection over Union (IoU). This accounts for slight discrepancies between human annotations while still rewarding correct grounding:
$$\mathcal{R}_G(o) = \mathrm{IoU}^{\times\eta} = \min\left(1, \eta \dfrac{\left|R(o) \cup R_y\right|}{\left|R(o) \cap R_y\right|}\right),$$
where $R(o)$ is the region extraction function that parses textual output $o$ to bounding boxes, $R_y$ is the annotated region, and $\eta$ is a relaxing constant.
This relaxation reward ensures full credit to the model when the regions annotated by models are in good correlation with human-annotated ones.

\paragraph{Output format validity (\textit{Format}).}
To ensure the model understands the structural requirements of the task, we introduce a format reward that encourages outputs conforming to the expected syntax. A valid output must include correctly structured \verb|<think>|, \verb|<tag>| and \verb|<verdict>| markers, as well as bounding boxes and captions that are syntactically well-formed and can be parsed using regular expressions.
Formally, the reward is defined as:
$$\mathcal{R}_F(o) = \begin{cases} 1, \quad & \text{if } V(o), R(o), T(o), C(o) \text{ are all parsable,} \\ -1, \quad & \text{o.w.} \end{cases}$$
where $T(o)$ and $C(o)$ extracts the caption of regions and image-level tags from $o$.

\subsection{RLHF stages with group relative policy optimization}
\begin{figure}[t]
\centering
\begin{minipage}{0.38\textwidth}
  \centering
  \captionof{table}{Weights for different GRPO stages used in the training pipeline.}
  \resizebox{\textwidth}{!}{%
    \input{tables/method-grpo-weights}
  }
  \label{tab:method-grpo-weights}
  
  \vspace{1em}
  
  \includegraphics[width=\textwidth,keepaspectratio=true]{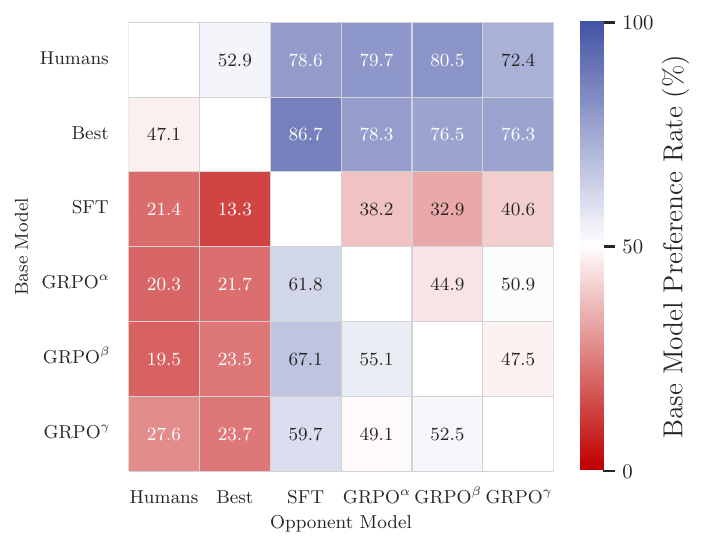}
  \captionof{figure}{Human preference matrix.}
  \label{fig:preference-matrix}
\end{minipage}
\hfill
\begin{minipage}{0.60\textwidth}
  \centering
  \includegraphics[width=\textwidth,keepaspectratio=true]{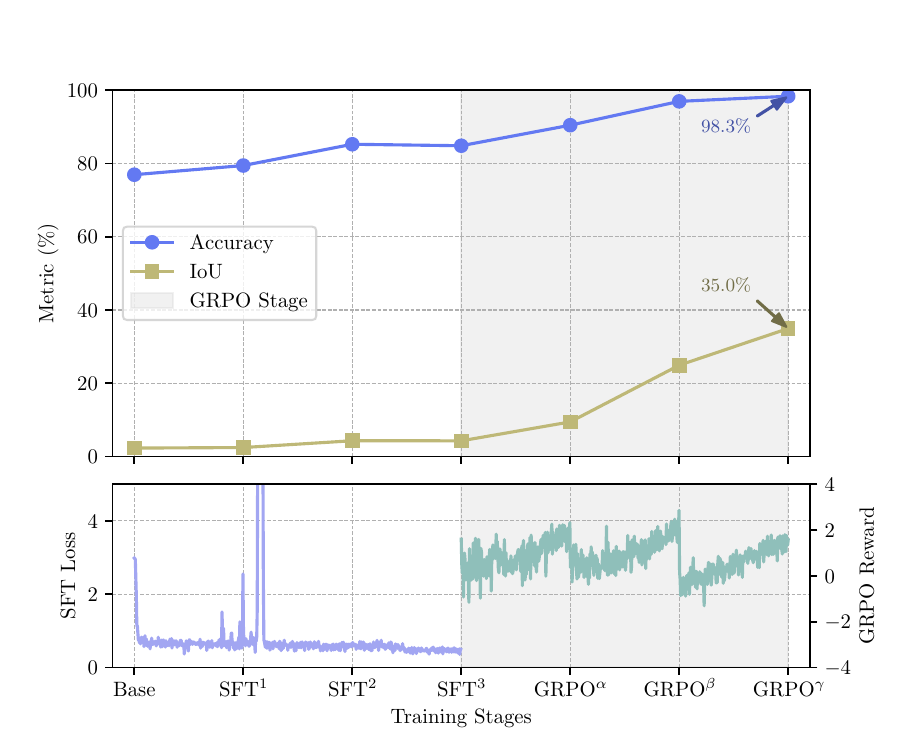}
  \caption{Accuracy, IoU metric (upper), loss and reward curves (lower) of the model during the training process.}
  \label{fig:training-acc}
\end{minipage}
\end{figure}

After SFT, the model will continue to go through three different GRPO stages with variably-weighted reward functions. GRPO progressively aligns the MLLM with our objectives of interpretable and reliable fake image detection by combining structured supervision from the dataset with targeted reward signals.
The final reward function can be written as $\mathcal{R} = r_\text{base} + \omega_G\mathcal{R}_G+\omega_C\mathcal{R}_C+\omega_F\mathcal{R}_F$, where $r_\text{base}$ is a stage-dependent base reward that ensures stable training dynamics across stages. $(r_\text{base}, \omega_G, \omega_C, \omega_R)$ varies with stage; their exact values are provided in \cref{tab:method-grpo-weights}.
It is worth noting that for $\mathcal{R}_C$ (Label) and $\mathcal{R}_F$ (Format), we may apply asymmetric weighting for positive and negative outcomes (\textit{e.g.}, +2/–1), allowing finer control over the reinforcement signal strength.

\paragraph{Stage \(\alpha\).} The first stage prioritizes correct formatting by assigning it a higher reward weight. It maintains a balanced weighting for IoU and label accuracy (using +1/-1 rewards). This phase aims to solidify the model's ability to adhere to the required output structure established during SFT.

\paragraph{Stage \(\beta\).} This phase shifts focus towards improving the model's core detection and localization performance. The reward weighting for IoU is increased (1.5x), and the reward for correct labeling is doubled. This stage penalizes wrong predictions and wrong formats, focusing on improving the detection capabilities of the model.

\paragraph{Stage \(\gamma\).} This stage further refines the localization accuracy by increasing the IoU reward weight (2.0x). The rewards for labeling and formatting are adjusted again (both 0.5/-1) to maintain balance.
In this stage, giving a correct prediction of label and format will lead to a reward of 0, which consequently challenges the MLLM's visual grounding capabilities to identify the actual artifacts present in the images.

%% file: tables/method-grpo-weights.tex
\begin{tabular}{@{}lccc@{}}
  \toprule
  GRPO Stages & $\alpha$ & $\beta$ & $\gamma$ \\ \midrule
  Base & 0 & -0.5 & -1 \\ \midrule
  IoU$^{\times 1.1}$ $\mathcal{R}_G$ & $1.0 \times$ & $1.5 \times$ & $2.0 \times$ \\ \midrule
  Label $\mathcal{R}_C$ & $+1$ / $-1$ & $+2$ / $-2$ & $0.5$ / $-1$ \\ \midrule
  Format $\mathcal{R}_F$ & $+2$ / $-1$ & $+1$ / $-1.5$ & $0.5$ / $-1$ \\ \bottomrule
\end{tabular}

%% file: sections/05-experiments.tex
\section{Experiments}
\label{sec:experiments}

\subsection{Experimental setup}
We adopt \href{https://huggingface.co/Qwen/Qwen2.5-VL-32B-Instruct}{Qwen2.5-VL-32B-Instruct}~\cite{Qwen25VL,Qwen2VL} as our base MLLM due to its balance between model size and performance, as well as its strong pre-trained grounding capabilities.
We trained our MLLMs on 8x NVIDIA A100 GPUs. All baseline methods are trained on one NVIDIA A100 GPU. The SFT period runs for three epochs, during which the training loss converges; each GRPO stage lasts one epoch.
More experimental details are provided in the \textit{supplementary materials}.

For baseline comparisons, we use the same training data as the MLLM setup. We adapt SegFormer~\cite{SegFormer} and ObjectFormer~\cite{ObjectFormer} to a multitask setting using our dataset.
Bounding box annotations are converted into binary masks for segmentation supervision, with real/synthetic classification employed as a parallel classification task.
For classification-only methods, including NPR~\cite{NPR}, DMD~\cite{DMimageDetection}, ComFor~\cite{ComFor}, AfPr~\cite{AntifakePrompt}, and DIRE~\cite{DIRE}, only image-level labels are used during training.

\input{tables/result-model-wise-acc-iou}
\input{tables/result-ood}
\begin{figure}[t]
    \centering
    \includegraphics[width=0.95\linewidth]{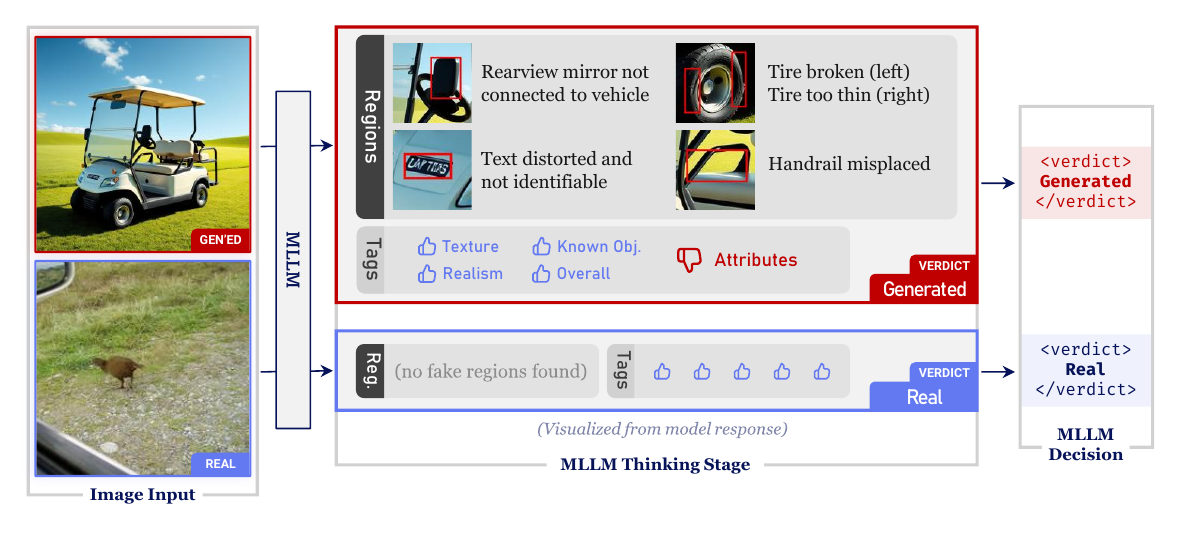}
    \caption{A sample user query and the corresponding model output (visualized).}
    \label{fig:demo}
\end{figure}

\subsection{Overall performance}
To ensure robustness and mitigate dataset bias, all models are evaluated using four-fold cross-validation. During training, the detection model is exposed to 75\% of images from the \textbf{FakeXplained} dataset along with an equal number of real samples. Evaluation is conducted on the remaining 25\% of synthetic images, again paired with the same number of real images.
We report both classification accuracy and localization performance using the IoU metric. Note that IoU is computed only on synthetic images, as real images do not include annotated fake regions.

\paragraph{Effects of different training stages}
To analyze the impact of each training stage, we report both accuracy and IoU metrics across the entire training process in \cref{fig:training-acc}. 
Without GRPO, SFT alone yields only marginal improvements over the base model.
As the training progresses, each GRPO stage strategically adjusts the reward weights to highlight different aspects of model behavior.
Notably, GRPO$^\beta$ introduces a significant boost in both accuracy and IoU, owing to the increased emphasis on classification ($\omega_C$) and grounding ($\omega_G$) rewards. 
The model reaches its best overall performance after completing GRPO$^\gamma$, demonstrating the cumulative benefit of the progressive reward design.

\paragraph{Human preference evaluation.} 
While IoU and classification accuracy provide objective metrics for detection performance, they do not fully capture the qualitative aspects of region-caption alignment. In fact, the model may, in some instances, generate annotations that surpass those of the original human annotators. To comprehensively assess the quality and relevance of the generated explanations, we conducted a human preference study involving an independent group of evaluators.
In this study, participants were shown pairs of outputs for the same image, each with different bounding box annotations and associated captions. No tags or metadata were provided. Evaluators were asked to choose the annotation that demonstrated better alignment between the region and caption, as well as higher overall annotation quality. If no clear preference emerged, a neutral option was available.

After removing neutral responses, we obtained 1,525 valid preference votes. The comparative results are shown in \cref{fig:preference-matrix}. The ``Humans'' category represents annotations from the \textbf{FakeXplained} dataset. When compared directly, human annotations were preferred in 52.9\% of cases, indicating that our model achieves near-parity with human annotators in producing region-grounded explanations, demonstrating the effectiveness of our framework in generating high-quality and explainable visual-textual reasoning.

\paragraph{Qualitative examples.}
\Cref{fig:demo} presents two representative examples of the model's output. In the upper example, the MLLM successfully identifies anomalous visual attributes, accurately localizes the flawed regions, and generates a coherent explanation, concluding with a correct verdict that the image is AI-generated. In the lower example, the model finds no obvious errors or artifacts in the image, reaching a verdict that the image is real. Additional qualitative examples will be provided in the \textit{supplementary materials}.

\paragraph{Comparing to other methods.}
Quantitative results are presented in \cref{tab:result-model-wise-acc-iou}. Our best-performing model achieves an overall classification accuracy of \textbf{98.1\%}, demonstrating strong robustness and consistent performance across different image generators.
For localization, the model achieves an IoU score of \textbf{37.8\%}, outperforming all segmentation-based baselines, including ObjectFormer~\cite{ObjectFormer} and SegFormer~\cite{SegFormer}.
This highlights the model’s ability to identify fake regions that are more consistent with human annotations.

\paragraph{Out-of-distribution (OoD) evaluation.} We also evaluated the models on two OoD datasets, FaceForensics++~\cite{FaceForensicsPP} and images generated by ChatGPT-4o~\cite{OpenAI4oGen, Dataset4oPreference}. As shown in \cref{tab:result-ood}, our model consistently outperforms all other methods across both datasets, demonstrating superior generalization to unseen generation domains.

\input{tables/result-ablations}

\subsection{Ablation studies}

Each variant in \cref{tab:ablations} isolates one training component to assess its individual contribution to overall model performance.

\paragraph{Without FakeXplained/fine-tuning.}
Without any fine-tuning on FakeXplained (\textit{No-FT}), \textit{Qwen-2.5-VL-32B-Instruct} achieves a modest accuracy of 73.4\%, indicating inadequate task-specific performance in its original state. The SFT stage alone improves the accuracy to 89.3\%, and incorporating RLHF further improves it to 98.1\%.
These results demonstrate the substantial performance gains enabled by our two-stage fine-tuning pipeline and underscore the necessity of the \textbf{FakeXplained} dataset for enabling explainable and reliable fake image detection.

\paragraph{Partial data.}

Our dataset consists of three essential components: image-level tags, region-level annotations with bounding boxes and captions, and binary real/fake labels. We conduct ablations by removing each component in turn to evaluate its impact.
All ablation variants are trained using our complete two-stage pipeline (SFT + RLHF). If a component is missing, it is excluded from both the SFT phase and the corresponding reward terms in RLHF. Specifically, when bounding boxes are unavailable, the $\mathcal{R}_G$ reward is disabled, while other rewards remain active.

Using only binary labels (\textit{label-only}) results in the lowest accuracy among partial data variants at 93.5\%.  This highlights the importance of structured reasoning information (\textit{i.e.}, tags, regions, and captions) in improving detection accuracy.
Nonetheless, this score is still considerably higher than the no-fine-tuning baseline (73.4\%) and also surpasses the best image-only detector DMD (92.8\%), showcasing the strong adaptability of MLLMs with two-stage fine-tuning for this task.

When bounding box annotations and the corresponding captions are removed (\textit{no-bbox}), the accuracy dropped by 2.5\%. IoU is not computed in this setting. This performance degradation demonstrates the value of grounding supervision during training, which helps the model focus attention on semantically meaningful regions and enhances its decision-making process.

Removing captions (\textit{no-caption}) results in a slight accuracy drop similar to the \textit{no-bbox} variant, but causes a significant decrease in IoU to 26.1\%. This indicates that textual descriptions play a crucial role in supporting accurate region-level localization. Captions also help the model articulate why a region is fake, thus improving both interpretability and grounding precision.

Excluding tags (\textit{no-tags}) yields a marginal drop in both the classification accuracy and the IoU. Tags capture perceptual attributes of the images, and their removal slightly degrades the model's holistic understanding of image quality, though the impact is less severe than removing bounding boxes or captions.

\paragraph{Training strategies.}
We evaluate the importance of dynamic reward weighting by comparing our progressive GRPO strategy against fixed-weight configurations (GRPO$^\alpha$, GRPO$^\beta$, and GRPO$^\gamma$) with the same total number of training steps. 
As shown in the corresponding columns, all fixed-weight variants underperform compared to our full progressive GRPO pipeline.
Notably, GRPO$^\gamma$ produces a higher IoU than our best model due to its specialized reward design, but this comes at the expense of classification accuracy. \Cref{fig:preference-matrix} also shows that these models are inferior to the best model, receiving no more than 30\% of non-neutral votes when compared to human-annotated data.

The complete pipeline delivers the most balanced performance across all evaluation criteria, demonstrating that dynamic weighting of rewards is essential for simultaneously optimizing output formatting, classification accuracy, localization precision, and region-caption correlation.

%% file: tables/result-model-wise-acc-iou.tex
\begin{table}[t]
\centering
\caption{Experimental result for current AI-generated image detectors and our MLLM-based method across different image generation methods.}
\label{tab:result-model-wise-acc-iou}
\resizebox{\columnwidth}{!}{

\begin{tabular}{@{}ccccccccccccc@{}}
\toprule
                             & \multicolumn{2}{c}{Ours} & \multicolumn{2}{c}{ObjectFormer~\cite{ObjectFormer}} & \multicolumn{2}{c}{SegFormer~\cite{SegFormer}} & NPR~\cite{NPR}   & DMD.~\cite{DMimageDetection}            & ComFor.~\cite{ComFor}         & AfPr.~\cite{AntifakePrompt}  & AEROB.~\cite{AEROBLADE} & DIRE~\cite{DIRE}  \\ \cmidrule(l){2-13} 
\multirow{-2}{*}{Generators} & Acc.            & IoU    & Acc.            & IoU            & Acc.               & IoU      & Acc.  & Acc.           & Acc.           & Acc.  & Acc.      & Acc.  \\ \midrule
DALL·E 2~\cite{DALLE2}                    & \textbf{0.986}  & 0.360  & 0.957           & 0.251          & 0.942              & 0.285    & 0.907 & 0.934          & 0.877          & 0.892 & 0.823     & 0.916 \\
DALL·E 3~\cite{DALLE3}                     & \textbf{0.991}  & 0.365  & 0.949           & 0.258          & 0.950              & 0.292    & 0.912 & 0.942          & 0.872          & 0.907 & 0.821     & 0.923 \\
DDIM~\cite{DDIM}                         & \textbf{0.974}  & 0.345  & 0.954           & 0.285          & 0.945              & 0.280    & 0.917 & 0.928          & 0.879          & 0.915 & 0.839     & 0.912 \\
DDPM~\cite{DDPM}                         & \textbf{0.979}  & 0.350  & 0.951           & 0.293          & 0.947              & 0.288    & 0.903 & 0.931          & 0.876          & 0.898 & 0.836     & 0.917 \\
FLUX.1-dev~\cite{FLUX}                   & \textbf{0.988}  & 0.362  & 0.958           & 0.299          & 0.940              & 0.295    & 0.922 & 0.937          & 0.874          & 0.779 & 0.843     & 0.919 \\
FLUX.1-schnell               & \textbf{0.972}  & 0.343  & 0.953           & 0.287          & 0.943              & 0.283    & 0.926 & 0.929          & 0.882          & 0.805 & 0.827     & 0.913 \\
GLIDE~\cite{GLIDE}                        & \textbf{0.970}  & 0.340  & 0.950           & 0.289          & 0.946              & 0.286    & 0.913 & 0.935          & 0.873          & 0.661 & 0.822     & 0.922 \\
Midjourney v4~\cite{Midjourney}                & \textbf{0.990}  & 0.364  & 0.956           & 0.296          & 0.949              & 0.294    & 0.908 & 0.939          & 0.869          & 0.878 & 0.814     & 0.925 \\
Midjourney v5                & \textbf{0.992}  & 0.366  & 0.959           & 0.273          & 0.941              & 0.297    & 0.902 & 0.943          & 0.871          & 0.851 & 0.718     & 0.927 \\
SD 1.4~\cite{SD14}                       & 0.968           & 0.338  & 0.952           & 0.286          & 0.944              & 0.282    & 0.921 & \textbf{0.970} & 0.880          & 0.852 & 0.951     & 0.909 \\
SD 1.5                       & \textbf{0.975}  & 0.347  & 0.955           & 0.294          & 0.948              & 0.290    & 0.916 & 0.949          & 0.875          & 0.866 & 0.966     & 0.915 \\
SD 2.1~\cite{SD21}                       & \textbf{0.980}  & 0.352  & 0.951           & 0.291          & 0.942              & 0.287    & 0.911 & 0.938          & 0.872          & 0.881 & 0.833     & 0.918 \\
SD 3.5 Large~\cite{SD3}                 & \textbf{0.991}  & 0.365  & 0.954           & 0.294          & 0.945              & 0.293    & 0.904 & 0.944          & 0.870          & 0.934 & 0.830     & 0.924 \\
SD 3.5 Large Turbo           & \textbf{0.993}  & 0.368  & 0.957           & 0.312          & 0.950              & 0.296    & 0.906 & 0.947          & 0.868          & 0.927 & 0.837     & 0.928 \\
VQDM~\cite{VQDM}                         & \textbf{0.973}  & 0.342  & 0.953           & 0.288          & 0.943              & 0.284    & 0.927 & 0.932          & 0.877          & 0.938 & 0.932     & 0.914 \\
\rowcolor[HTML]{DCE2F2} 
\textit{Diffusion} & \textbf{0.983} & 0.356 & 0.954 & 0.287 & 0.945 & 0.290 & 0.913 & 0.941 & 0.874 & 0.864 & 0.842 & 0.920 \\ \midrule
BigGAN~\cite{BigGAN}                       & \textbf{0.965}  & 0.335  & 0.950           & 0.280          & 0.941              & 0.278    & 0.918 & 0.892          & 0.903          & 0.933 & 0.861     & 0.887 \\
GALIP~\cite{GALIP}                        & 0.882           & 0.353  & \textbf{0.941}           & 0.279          & \textbf{0.941}     & 0.289    & 0.882 & 0.882          & 0.941          & 0.353 & 0.706     & 0.882 \\
VQGAN~\cite{VQGAN}                        & \textbf{0.967}  & 0.337  & 0.954           & 0.282          & 0.943              & 0.280    & 0.907 & 0.889          & 0.908          & 0.921 & 0.932     & 0.885 \\
StyleGAN-XL~\cite{StyleGAN}                  & 0.960           & 0.330  & 0.951           & 0.278          & 0.940              & 0.275    & 0.914 & 0.884          & \textbf{0.980} & 0.928 & 0.939     & 0.879 \\
\rowcolor[HTML]{DCE2F2} 
\textit{GAN} & \textbf{0.955} & 0.337 & 0.950 & 0.280 & 0.941 & 0.279 & 0.912 & 0.890 & 0.916 & 0.866 & 0.860 & 0.885 \\ \midrule
PixArtAlpha~\cite{PixArtAlpha}                  & \textbf{0.987}  & 0.357  & 0.956           & 0.295          & 0.947              & 0.291    & 0.908 & 0.912          & 0.891          & 0.934 & 0.927     & 0.903 \\
PixArtDelta~\cite{PixArtDelta}                  & \textbf{0.984}  & 0.354  & 0.953           & 0.292          & 0.943              & 0.289    & 0.921 & 0.909          & 0.893          & 0.939 & 0.922     & 0.899 \\
PixArtSigma~\cite{PixArtSigma}                  & \textbf{0.989}  & 0.360  & 0.957           & 0.296          & 0.949              & 0.293    & 0.919 & 0.915          & 0.889          & 0.924 & 0.931     & 0.905 \\
DiT~\cite{DiT}                          & \textbf{0.978}  & 0.349  & 0.952           & 0.290          & 0.942              & 0.287    & 0.913 & 0.907          & 0.896          & 0.928 & 0.938     & 0.897 \\
\rowcolor[HTML]{DCE2F2} 
\textit{DiT} & \textbf{0.983} & 0.354 & 0.954 & 0.293 & 0.945 & 0.289 & 0.914 & 0.910 & 0.893 & 0.931 & 0.931 & 0.900 \\ \midrule
VAR~\cite{VAR}                          & \textbf{0.976}  & 0.346  & 0.954           & 0.287          & 0.945              & 0.283    & 0.928 & 0.893          & 0.901          & 0.934 & 0.927     & 0.889 \\
Infinity~\cite{Infinity}                     & \textbf{0.974}  & 0.344  & 0.951           & 0.289          & 0.941              & 0.286    & 0.914 & 0.897          & 0.899          & 0.938 & 0.924     & 0.883 \\
MaskGIT~\cite{MaskGIT}                      & \textbf{0.972}  & 0.342  & 0.955           & 0.288          & 0.948              & 0.284    & 0.909 & 0.895          & 0.904          & 0.923 & 0.933     & 0.886 \\
LlamaGen~\cite{LlamaGen}                     & \textbf{0.980}  & 0.351  & 0.953           & 0.429          & 0.944              & 0.289    & 0.923 & 0.899          & 0.897          & 0.929 & 0.938     & 0.892 \\
\rowcolor[HTML]{DCE2F2} 
\textit{Others} & \textbf{0.978} & 0.348 & 0.953 & 0.369 & 0.944 & 0.287 & 0.920 & 0.898 & 0.898 & 0.931 & 0.933 & 0.889 \\ \midrule
Real Images~\cite{ImageNet}                  & \textbf{0.985}  & -  & 0.956           & -          & 0.946              & -    & 0.918 & 0.903          & 0.882          & 0.934 & 0.854     & 0.896 \\ \midrule
\rowcolor[HTML]{DCE2F2} 
Overall & \textbf{0.981} & 0.352 & 0.954 & 0.299 & 0.945 & 0.289 & 0.914 & 0.928 & 0.882 & 0.887 & 0.873 & 0.911 \\ \bottomrule
\end{tabular}

}
\end{table}

%% file: tables/result-ood.tex
\begin{table}[t]
\centering
\caption{The accuracy of different detection methods on unseen image categories. All enlisted methods are trained on FakeXplained.}
\label{tab:result-ood}
\resizebox{\columnwidth}{!}{%
\begin{tabular}{@{}cccccccccc@{}}
\toprule
Sources                            & \textbf{Ours}  & ObjectFormer & SegFormer & NPR         & DMD   & ComFor & AfPr  & AEROBLADE & DIRE        \\ \midrule
OpenAI 4o Preference~\cite{Dataset4oPreference} & \textbf{0.803} & 0.513        & 0.538     & 0.790       & 0.735 & 0.636  & 0.597 & 0.458     & {\ul 0.793} \\
FaceForensics++~\cite{FaceForensicsPP}                   & \textbf{0.879} & 0.598        & 0.716     & {\ul 0.861} & 0.562 & 0.429  & 0.746 & 0.681     & 0.850       \\ \bottomrule
\end{tabular}%
}
\end{table}

%% file: tables/result-ablations.tex
\begin{table}[t]
\centering
\caption{Comparisons of model performance when trained with different configurations.}
\label{tab:ablations}
\resizebox{0.9\columnwidth}{!}{%

\begin{tabular}{@{}ccccccccccc@{}}
\toprule
\multirow{2}{*}{Metric} &
  \multirow{2}{*}{\textbf{Best}} &
  \multirow{2}{*}{\textbf{No-FT}} &
  \multicolumn{4}{c}{Partial Data} &
  \multicolumn{4}{c}{Training Strategy} \\ \cmidrule(l){4-11} 
 &
   &
   &
  no-bbox &
  no-caption &
  no-tags &
  \multicolumn{1}{c|}{label-only} &
  SFT &
  GRPO$^\alpha$ &
  GRPO$^\beta$ &
  GRPO$^\gamma$ \\ \midrule
Acc. &
  \textbf{0.981} &
  0.734 &
  0.956 &
  0.958 &
  0.977 &
  0.935 &
  0.893 &
  0.954 &
  0.966 &
  0.947 \\
IoU &
  \textbf{0.378} &
  - &
  - &
  0.261 &
  0.375 &
  - &
  0.043 &
  0.309 &
  0.296 &
  0.401 \\ \bottomrule
\end{tabular}%
}


\end{table}

%% file: sections/07-conclusion.tex
\section{Conclusion}
\label{sec:conclusion}
Our research presents an explainable AI-generated image detection approach utilizing MLLMs that transcends binary classification by providing human-interpretable explanations alongside accurate detection results. Through a novel progressive GRPO paradigm, the system achieves superior performance metrics ($98.1\%$ accuracy, $37.8\%$ IoU, sound human preference) compared to conventional methods. The work addresses the critical need for transparent detection systems that augment human judgment in an era of advancing generative technologies, establishing a foundation for explainable visual media authentication that articulates the rationale behind algorithmic decisions.

%% file: sections/suppl.tex
\section*{Appendix}

\section{Additional experimental details}

\subsection{Chat templates}

Annotated images are preprocessed into conversational formats before being used to train MLLMs.
A chat template specifies how the input components are organized to generate a trainable conversation $\mathcal{T}_i$. Formally:

$$\mathbb{T}(\mathtt{image}, \mathtt{regions}, \mathtt{captions}, \mathtt{tags}, \mathtt{is\_real})\to\mathcal{T}_i,$$

\begin{figure}[t]
    \centering
    \includegraphics[width=1\linewidth]{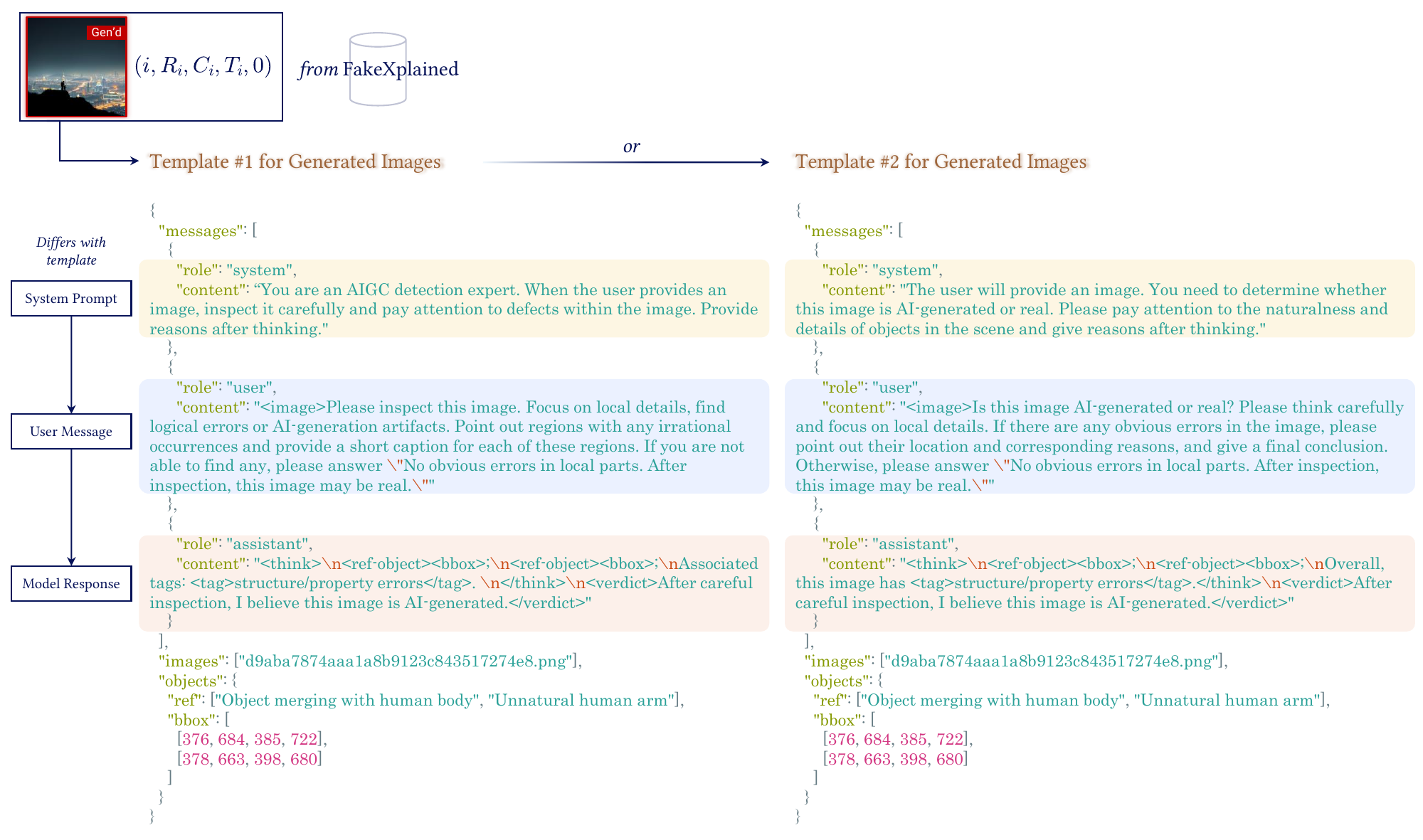}
    \caption{An example showing two different chat templates branched from one annotation entry.}
    \label{fig:chat-templates-showcase}
\end{figure}

To address potential overfitting, we developed nine distinct chat templates: three for real images and six for generated images, randomly selected during pre-processing.
\Cref{fig:chat-templates-showcase} illustrates sample conversations derived from a single annotated image using different templates.
Although system prompts and user queries vary, the model responses do not change significantly in order to maintain a consistent structure, ensuring the output parser remains independent of the user query.
The impact of the number of chat templates is discussed in~\cref{subsec:chat-templates}.

\subsection{Training details}

\paragraph{Two-stage training.}
We use \href{https://github.com/modelscope/ms-swift/}{ms-swift}~\cite{ms-swift} for fine-tuning Qwen-2.5-VL models.

In the LoRA SFT stage, we noticed that freezing either the projector or the vision encoder leads to marginal improvement over the base model without training. To achieve optimal SFT performance, both modules must be fine-tuned jointly. 

After the SFT stage, we use GRPO instead of PPO.
As noted in~\cite{DeepSeekMath}, GRPO obviates the need for additional value function approximation as in PPO, and instead uses the average reward of multiple sampled outputs.
For each query $q$, GRPO samples $G$ outputs $\{o_1, o_2, \ldots, o_G\}$ from the old policy model $\pi_{\theta_{old}}$, and uses the relative \textit{advantage} to optimize the MLLM, making it particularly well-suited for multi-modal reasoning tasks where absolute reward calibration is challenging.

We set the initial learning rate to \( 10^{-4} \) for the SFT stage and \( 10^{-5} \) for the RLHF stage.
As shown in Figure 3 of the main paper, despite a few loss spikes occurring at random steps during the SFT training process, the whole SFT process is relatively stable.
Reward signals fluctuated early in each GRPO stage but quickly converged as the model adapted, confirming the effectiveness of our reward design and training strategy.

\paragraph{Computational resources.}
The full training procedure took 41.0 hours on 8x NVIDIA A100 (80G) GPUs, among which 16.5 hours (40.2\%) were spent on the SFT stage.

At inference time, the end-to-end pipeline that takes an image as input, giving a verdict and grounding information (if the image is deemed AI-generated) takes an average of 7.8 seconds on 2x NVIDIA A100 (80G) GPUs.

\section{Annotation details of FakeXplained}

\paragraph{Annotator qualification and training.}
All recruited annotators have prior experience in photography, have seen AI-generated images before, possess a fundamental photographic literacy understanding, and are familiar with related concepts such as ``saturation,'' ``shadow,'' ``perspective,'' and ``noise.''

\paragraph{Instructions on fake regions.} 
The rule for annotating fake regions is, if through observation of the selected regions of interest, humans should be able to clearly determine that the image is not an authentic photograph.
Fake regions primarily show objects that do not follow the natural physics laws, or contradict common sense.
Common image generation artifacts are also encouraged to be annotated.
After selecting a local area in the image, it is necessary to describe the reason for identifying it as a generated image.
The descriptive sentence must start with a noun, followed by one or several adjective phrases or short clauses, and must exclusively describe content that appears in the region.

\paragraph{Definition of tags.}
We refer to the most prominent depicted object in the non-background portion of the image as the \textit{image subject}.
There are exactly five different tags that annotators can attach to an image.
Their definitions are listed below.

\begin{itemize}[leftmargin=*]
\item \textbf{Perspective errors:} Indicates that the image has an unnatural viewing angle, or errors in perspective, vanishing points. Incorrect occlusion and shadow errors do not constitute perspective errors, but can be considered as fake regions instead.
\item \textbf{Artistic styles:} If the overall image presents any artistic style, including but not limited to oil painting, ink painting, or manga style, then select the ``Artistic Style'' tag. If only a certain part of the image contains content in an artistic style, this tag should not be selected.
\item \textbf{Unknown objects:} Indicates that the \textit{subject} of the image does not exist in the world, or is obviously unreasonable. There may be unusual insects and furniture with strong design elements. Judgment should be based on intuition; unfamiliar or rare subjects do not necessarily indicate unreasonable or non-existent objects.
\item \textbf{Structure/attribute errors:} Indicates that the \textbf{subject} of the image has a structure that is inconsistent with common knowledge, or has attributes inconsistent with common knowledge. Examples include green flower petals, pink elephants, bent iron spoon handles, humans with more than two legs, and asymmetrical shapes. For erroneous attributes that only occupy a small portion of the image subject, such as an incorrect number of fingers on a human hand, fake regions should be marked as well.
\item \textbf{Texture errors:} If obvious texture errors appear in the image, this tag needs to be selected. For example, the texture of the entire image is blurry, or a portion of an object has a repetitive, tilted, or distorted texture. Unreadable text does not qualify as a texture error and should be labeled as a fake region instead. If ``Artistic Style'' has already been marked, this tag is usually omitted.
\item \textbf{Other anomalies:} If there are very obvious global errors in the image that do not belong to any of the above categories, check this item. This tag can also be marked even if other tags have already been chosen.
\end{itemize}

\section{More examples}

\begin{figure}[p]
    \centering
    \includegraphics[width=1\linewidth]{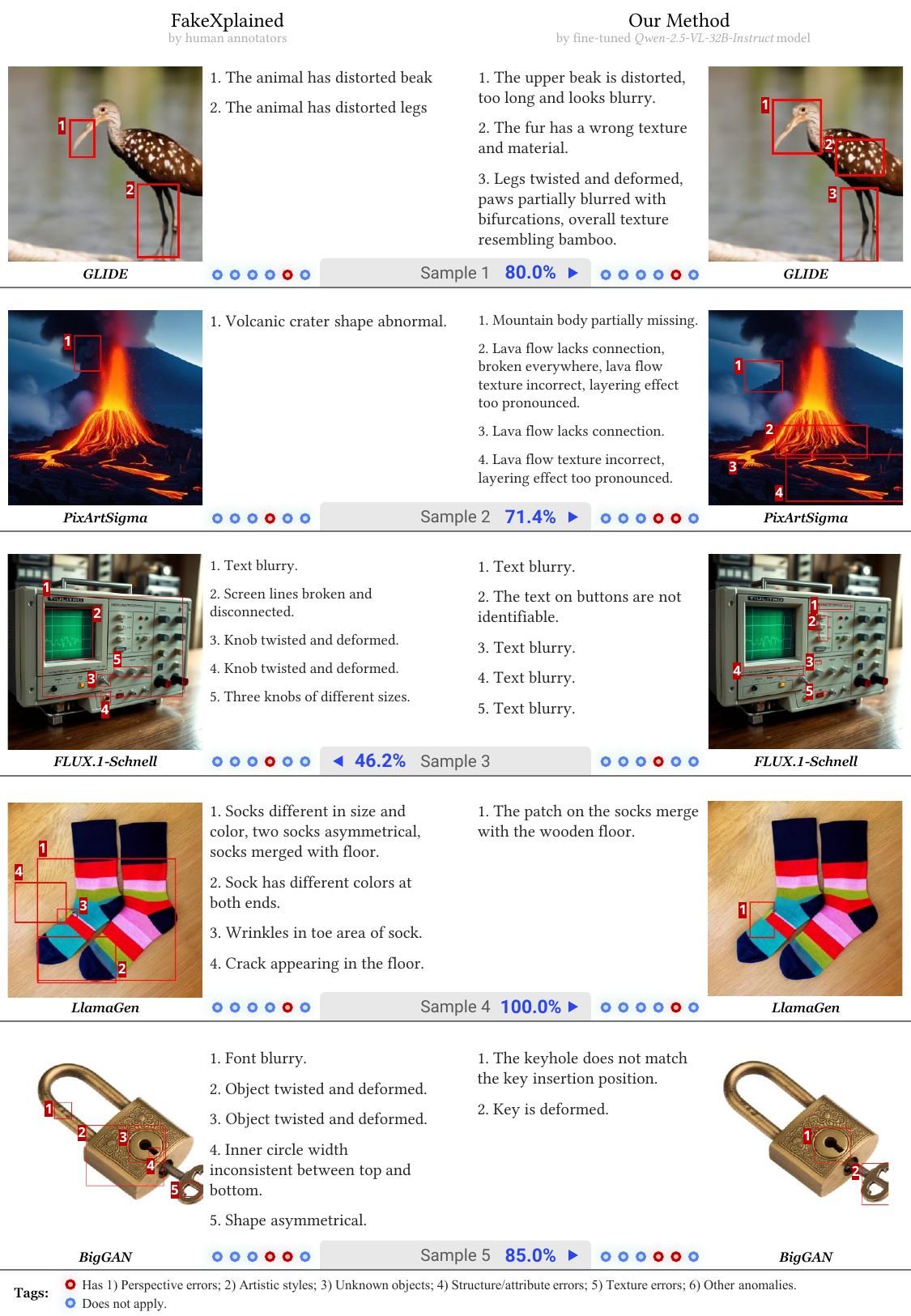}
    \caption{More annotation examples from FakeXplained and model response visualized from the fine-tuned MLLM. The ratio in the center shows the human preference score.}
    \label{fig:more-examples}
\end{figure}

\Cref{fig:more-examples} presents more annotated AI-generated images from FakeXplained.
The left column displays the human annotations of FakeXplained.
The right column shows the inference results of our best model.
The center bar indicates the proportion of human preference votes from our user study.
Note that since the ``neutral'' option was allowed, although the third annotated image received 46.2\% of the votes, the human annotator is still rated higher than our model response.
Our model demonstrates the ability to generate clearer, more descriptive captions for fake regions and reliably identifies content that contradicts common sense.
For instance, in the lock-and-keyhole example (row 5), the model successfully detects that the key is not inserted into the correct keyhole. 
In the volcano example (row 2), in addition to identifying the ``broken mountain body'' as in the human annotation, the model also detects a subtle issue: the disconnection of the lava flow, highlighting its fine-grained visual reasoning capabilities.

\section{Robustness against image perturbations}

To evaluate the practical applicability of our approach, we conduct a comprehensive robustness evaluation under common image degradations that are frequently encountered in real-world scenarios.
\Cref{tab:result-degradation} presents a comparative performance analysis across three perturbation categories: JPEG compression, random cropping, and downsampling.

Our method demonstrates exceptional resilience to JPEG compression artifacts, achieving low performance degradations of merely $0.3\%$ and $0.8\%$ from the uncompressed baseline, significantly outperforming current state-of-the-art methods. all of which experience at least a $3\%$ degradation. Notably, SegFormer and ObjectFormer show more stability than image-only classification models, indicating that grounding enhances robustness, although they still fall short of our method.
For downsampling, we scaled the input images to 50\% of their original width and height. In random cropping and downsampling experiments, our approach maintains the original accuracy of $0.981$, indicating robust performance across different resolution scales. Meanwhile, downsampling does not severely affect the IoU score, which suggests that our grounded reasoning approach effectively captures semantic-level artifacts that remain detectable even at reduced resolutions, unlike methods that may rely on pixel-level features more susceptible to resolution changes.
Since random cropping modifies the overall image layout, this action can remove certain fake regions from an image entirely, leading to lower IoU across all methods.
Interestingly, we observe a slight increase in IoU after downsampling. We hypothesize that this is because our grounding model focuses on the dominant artifact region, which remains visible at lower resolutions, while noisy fine details are suppressed, leading to more precise and focused localization. 
Overall, the consistent performance across perturbation types demonstrates that our model captures underlying semantic artifacts in AI-generated content, enabling robust detection even under challenging image conditions.

\input{tables/result-degradation}

\section{Additional ablation studies}

\subsection{Out-of-distribution performance}

We evaluate generalization capabilities of the ablation models in Section 5.3 (Table 4) of the main paper on two OoD datasets: 4o~\cite{Dataset4oPreference} and FaceForensics++~\cite{FaceForensicsPP}. Table~\ref{tab:result-abl-ood} shows that our complete pipeline achieves accuracies of 0.803 and 0.879 respectively, compared to 0.412 and 0.565 for the base model without fine-tuning.

Among partial data ablations, the label-only configuration performs poorly, achieving accuracies similar to the non-fine-tuned baseline (45.6\% vs 41.2\% on 4o), confirming that spatial grounding is essential for generalization.

The SFT stage alone yields moderate performance (0.587 on 4o, 0.678 on FF++). Subsequent GRPO stages improve results, with GRPO$^\beta$ performing best among the variants. This result is consistent with  findings discussed in our main paper, as the RLHF stages give more performance boost than the SFT stage. The consistent improvements across both datasets suggest our approach learns generalizable features for AI-generated content detection rather than dataset-specific patterns.

\input{tables/result-abl-ood}

\subsection{Disabling LoRA}

We employ LoRA during training to reduce computational cost and memory usage. While full-parameter fine-tuning is technically possible, our results show that it does not improve accuracy or IoU. In fact, performance slightly degrades on the FakeXplained test set, likely due to the limited amount of annotated data (see ``No-LoRA''). 
This suggests that LoRA provides a more efficient and suitable training strategy under current data constraints. With significantly more training data, full fine-tuning may yield better results.

\subsection{Number of chat templates}
\label{subsec:chat-templates}

When some of these templates are absent from the training data, the detection accuracy does not change significantly, while the localization capabilities decline slightly.
However, when evaluating on the OoD dataset, a larger performance gap is found between models trained with fewer chat templates, with a 3.3\% accuracy drop for the 3+2 group and 2.7\% for the 1+1 group on the FaceForensics++~\cite{FaceForensicsPP} dataset, indicating that generalization capabilities can be improved with more chat templates.

\subsection{Model size variants}

We evaluate the 3B and 7B variants of Qwen-2.5-VL and observe that they behave quite differently from the 32B model.
When fine-tuning these lighter variants, both models show marginal performance improvement with their corresponding base models.
These results confirm that small models lack the capacity to support the complex visual reasoning and grounding required for this task, reinforcing the necessity of large-scale models.

\subsection{Other base models}

While InternVL-2.5~\cite{InternVL25} also supports visual grounding, when trained using the same pipeline, InternVL-2.5 still suffers from hallucination issues, producing a non-negligible proportion (13.7\%) of responses that cannot be parsed. This occurs primarily due to the absence of grounding information in the responses, despite fine-tuning and carefully designed user prompts intended to guide the model toward providing such information.

\input{tables/result-more-ablations}

\section{Limitations \& future works}
Despite promising results, our approach still has limitations. The Qwen-2.5-VL-32B-Instruct model incurs substantial computational costs, which may limit deployment in resource-constrained environments. 
Our evaluation does not sufficiently cover domain-specific or real-world image types, such as medical, industrial, or artistic imagery.
Future work should enhance robustness through domain-adaptive fine-tuning, data augmentation, and pre-processing pipelines that account for quality degradation and layout variations.

\section{Broader impact}
While our system improves interpretability in detecting AI-generated content, it may also introduce risks.
The detailed explanations of detection rationale could inadvertently assist malicious adversaries in developing more sophisticated evasion techniques, potentially contributing to an adversarial ``arms race.''
The deployment of such systems without careful consideration could lead to over-censorship of legitimate content, particularly affecting artists and creators who use AI tools ethically.
To mitigate these risks, we recommend responsible deployment frameworks, ongoing monitoring for bias and fairness, and collaborative development with stakeholders to ensure the technology serves the public interest while preserving legitimate creative expression.

%% file: tables/result-degradation.tex
\begin{table}[t]
\centering
\caption{Comparative performance analysis under compression artifacts, spatial transformations, and resolution changes.}
\label{tab:result-degradation}
\resizebox{\columnwidth}{!}{%
\begin{tabular}{@{}ccccccccccc@{}}
\toprule
\multicolumn{2}{c}{Degradation \& Metric} &
  \textbf{Ours} &
  ObjectF.~\cite{ObjectFormer} &
  SegF.~\cite{SegFormer} &
  NPR~\cite{NPR} &
  DMD.~\cite{DMimageDetection} &
  ComFor.~\cite{ComFor} &
  AfPr.~\cite{AntifakePrompt} &
  AEROB.~\cite{AEROBLADE} &
  DIRE~\cite{DIRE} \\ \midrule
\multirow{2}{*}{\begin{tabular}[c]{@{}c@{}}JPEG Compression\\ (80\% Quality)\end{tabular}} &
  Acc. &
  \textbf{0.978} &
  0.940 &
  0.927 &
  0.820 &
  0.908 &
  0.840 &
  0.871 &
  0.842 &
  0.884 \\
 &
  IoU &
  \textbf{0.350} &
  0.284 &
  0.231 &
  - &
  - &
  - &
  - &
  - &
  - \\ \midrule
\multirow{2}{*}{\begin{tabular}[c]{@{}c@{}}JPEG Compression\\ (30\% Quality)\end{tabular}} &
  Acc. &
  \textbf{0.973} &
  0.926 &
  0.915 &
  0.781 &
  0.897 &
  0.784 &
  0.856 &
  0.814 &
  0.879 \\
 &
  IoU &
  \textbf{0.322} &
  0.267 &
  0.198 &
  - &
  - &
  - &
  - &
  - &
  - \\ \midrule
\multirow{2}{*}{Random Cropping} &
  Acc. &
  \textbf{0.981} &
  0.943 &
  0.934 &
  0.903 &
  0.915 &
  0.829 &
  0.879 &
  0.858 &
  0.891 \\
 &
  IoU &
  \textbf{0.289} &
  0.217 &
  0.176 &
  - &
  - &
  - &
  - &
  - &
  - \\ \midrule
\multirow{2}{*}{\begin{tabular}[c]{@{}c@{}}Downsampling\\ (0.5x)\end{tabular}} &
  Acc. &
  \textbf{0.981} &
  0.929 &
  0.931 &
  0.899 &
  0.912 &
  0.853 &
  0.875 &
  0.841 &
  0.894 \\
 &
  IoU &
  \textbf{0.370} &
  0.259 &
  0.254 &
  - &
  - &
  - &
  - &
  - &
  - \\ \midrule
\multirow{2}{*}{\begin{tabular}[c]{@{}c@{}}Original \\ Images\end{tabular}} &
  Acc. &
  \textbf{0.981} &
  0.954 &
  0.945 &
  0.914 &
  0.928 &
  0.882 &
  0.887 &
  0.873 &
  0.911 \\
 &
  IoU &
  \textbf{0.352} &
  0.299 &
  0.289 &
  - &
  - &
  - &
  - &
  - &
  - \\ \bottomrule
\end{tabular}%
}
\end{table}

%% file: tables/result-abl-ood.tex
\begin{table}[t]
\centering
\caption{Out-of-distribution performance evaluation across different datasets when trained with various configurations mentioned in the paper.}
\label{tab:result-abl-ood}
\resizebox{\columnwidth}{!}{%
\begin{tabular}{@{}ccccccccccc@{}}
\toprule
\multirow{2}{*}{Dataset} & \multirow{2}{*}{\textbf{Best}} & \multirow{2}{*}{\textbf{No-FT}} & \multicolumn{4}{c}{Partial Data} & \multicolumn{4}{c}{Training Strategy} \\ \cmidrule(l){4-11} 
 &
   &
   &
  no-bbox &
  no-caption &
  no-tags &
  \multicolumn{1}{c|}{label-only} &
  SFT &
  GRPO$^\alpha$ &
  GRPO$^\beta$ &
  GRPO$^\gamma$ \\ \midrule
4o~\cite{Dataset4oPreference} &
  \textbf{0.803} &
  0.412 &
  0.765 &
  0.780 &
  0.797 &
  0.456 &
  0.587 &
  0.756 &
  0.788 &
  0.758 \\
FF++~\cite{FaceForensicsPP} &
  \textbf{0.879} &
  0.565 &
  0.834 &
  0.833 &
  0.831 &
  0.680 &
  0.678 &
  0.828 &
  0.831 &
  0.829 \\ \bottomrule
\end{tabular}%
}
\end{table}

%% file: tables/result-more-ablations.tex
\begin{table}[t]
\centering
\caption{More results from different training parameters, number of chat templates and models variants.}
\label{tab:result-more-ablations}
\resizebox{\columnwidth}{!}{%
\begin{tabular}{@{}cccccccccccccc@{}}
\toprule
\multirow{2}{*}{Dataset} &
  \multirow{2}{*}{Metric} &
  \multirow{2}{*}{\textbf{Best}} &
  \multirow{2}{*}{No-LoRA} &
  \multicolumn{2}{c}{Chat Templates} &
  \multicolumn{4}{c}{Model Size} &
  \multicolumn{4}{c}{InternVL 2.5} \\ \cmidrule(l){5-14} 
 &
   &
   &
   &
  1+1 &
  \multicolumn{1}{c|}{3+2} &
  3B (No-FT) &
  3B &
  7B (No-FT) &
  \multicolumn{1}{c|}{7B} &
  8B (No-FT) &
  8B &
  26B (No-FT) &
  26B \\ \midrule
\multirow{2}{*}{FakeXplained} &
  Acc. &
  \textbf{0.981} &
  0.976 &
  0.978 &
  0.982 &
  0.710 &
  0.715 &
  0.725 &
  0.823 &
  0.525 &
  0.762 &
  0.603 &
  0.805 \\
 &
  IoU &
  \textbf{0.378} &
  0.342 &
  0.374 &
  0.369 &
  - &
  0.127 &
  - &
  0.137 &
  - &
  0.121 &
  - &
  0.148 \\ \midrule
4o~\cite{Dataset4oPreference} &
  Acc. &
  0.803 &
  \textbf{0.806} &
  0.786 &
  0.795 &
  0.407 &
  0.625 &
  0.434 &
  0.690 &
  0.295 &
  0.385 &
  0.317 &
  0.581 \\
FF++~\cite{FaceForensicsPP} &
  Acc. &
  \textbf{0.879} &
  0.854 &
  0.858 &
  0.846 &
  0.488 &
  0.574 &
  0.595 &
  0.678 &
  0.410 &
  0.391 &
  0.412 &
  0.639 \\ \bottomrule
\end{tabular}%
}
\end{table}

%% file: neurips_2025.bbl
\begin{thebibliography}{10}

\bibitem{DefakeByReals}
Xiuli Bi, Bo~Liu, Fan Yang, Bin Xiao, Weisheng Li, Gao Huang, and Pamela~C.
  Cosman.
\newblock Detecting generated images by real images only, 2023.

\bibitem{BigGAN}
Andrew Brock, Jeff Donahue, and Karen Simonyan.
\newblock Large scale gan training for high fidelity natural image synthesis.
\newblock In {\em International Conference on Learning Representations}, 2018.

\bibitem{MaskGIT}
Huiwen Chang, Han Zhang, Lu~Jiang, Ce~Liu, and William~T Freeman.
\newblock Maskgit: Masked generative image transformer.
\newblock In {\em Proceedings of the IEEE/CVF conference on computer vision and
  pattern recognition}, pages 11315--11325, 2022.

\bibitem{AntifakePrompt}
You-Ming Chang, Chen Yeh, Wei-Chen Chiu, and Ning Yu.
\newblock Antifakeprompt: Prompt-tuned vision-language models are fake image
  detectors.
\newblock {\em arXiv preprint arXiv:2310.17419}, 2023.

\bibitem{PixArtSigma}
Junsong Chen, Chongjian Ge, Enze Xie, Yue Wu, Lewei Yao, Xiaozhe Ren, Zhongdao
  Wang, Ping Luo, Huchuan Lu, and Zhenguo Li.
\newblock Pixart-$\sigma$: Weak-to-strong training of diffusion transformer for
  4k text-to-image generation.
\newblock In {\em European Conference on Computer Vision}, pages 74--91.
  Springer, 2024.

\bibitem{PixArtDelta}
Junsong Chen, Yue Wu, Simian Luo, Enze Xie, Sayak Paul, Ping Luo, Hang Zhao,
  and Zhenguo Li.
\newblock Pixart-$\delta$: Fast and controllable image generation with latent
  consistency models.
\newblock {\em arXiv preprint arXiv:2401.05252}, 2024.

\bibitem{PixArtAlpha}
Junsong Chen, Jincheng Yu, Chongjian Ge, Lewei Yao, Enze Xie, Yue Wu, Zhongdao
  Wang, James Kwok, Ping Luo, Huchuan Lu, et~al.
\newblock Pixart-$\alpha$: Fast training of diffusion transformer for
  photorealistic text-to-image synthesis.
\newblock {\em arXiv preprint arXiv:2310.00426}, 2023.

\bibitem{R1-V}
Liang Chen, Lei Li, Haozhe Zhao, Yifan Song, and Vinci.
\newblock R1-v: Reinforcing super generalization ability in vision-language
  models with less than \$3.
\newblock \url{https://github.com/Deep-Agent/R1-V}, 2025.
\newblock Accessed: 2025-02-02.

\bibitem{InternVL25}
Zhe Chen, Weiyun Wang, Yue Cao, Yangzhou Liu, Zhangwei Gao, Erfei Cui, Jinguo
  Zhu, Shenglong Ye, Hao Tian, Zhaoyang Liu, Lixin Gu, Xuehui Wang, Qingyun Li,
  Yimin Ren, Zixuan Chen, Jiapeng Luo, Jiahao Wang, Tan Jiang, Bo~Wang, Conghui
  He, Botian Shi, Xingcheng Zhang, Han Lv, Yi~Wang, Wenqi Shao, Pei Chu,
  Zhongying Tu, Tong He, Zhiyong Wu, Huipeng Deng, Jiaye Ge, Kai Chen, Kaipeng
  Zhang, Limin Wang, Min Dou, Lewei Lu, Xizhou Zhu, Tong Lu, Dahua Lin,
  Yu~Qiao, Jifeng Dai, and Wenhai Wang.
\newblock Expanding performance boundaries of open-source multimodal models
  with model, data, and test-time scaling, 2025.

\bibitem{DMimageDetection}
Riccardo Corvi, Davide Cozzolino, Giada Zingarini, Giovanni Poggi, Koki Nagano,
  and Luisa Verdoliva.
\newblock On the detection of synthetic images generated by diffusion models.
\newblock In {\em IEEE International Conference on Acoustics, Speech and Signal
  Processing (ICASSP)}, pages 1--5, 2023.

\bibitem{ImageNet}
Jia Deng, Wei Dong, Richard Socher, Li-Jia Li, K.~Li, and Li~Fei-Fei.
\newblock Imagenet: A large-scale hierarchical image database.
\newblock {\em 2009 IEEE Conference on Computer Vision and Pattern
  Recognition}, pages 248--255, 2009.

\bibitem{ArtifactSpectrumDefake}
Tarik Dzanic, Karan Shah, and Freddie~D. Witherden.
\newblock Fourier spectrum discrepancies in deep network generated images.
\newblock In {\em Proceedings of the 34th International Conference on Neural
  Information Processing Systems}, NIPS '20, Red Hook, NY, USA, 2020. Curran
  Associates Inc.

\bibitem{SD3}
Patrick Esser, Sumith Kulal, Andreas Blattmann, Rahim Entezari, Jonas Müller,
  Harry Saini, Yam Levi, Dominik Lorenz, Axel Sauer, Frederic Boesel, Dustin
  Podell, Tim Dockhorn, Zion English, Kyle Lacey, Alex Goodwin, Yannik Marek,
  and Robin Rombach.
\newblock Scaling rectified flow transformers for high-resolution image
  synthesis, 2024.

\bibitem{VQGAN}
Patrick Esser, Robin Rombach, and Bjorn Ommer.
\newblock Taming transformers for high-resolution image synthesis.
\newblock In {\em Proceedings of the IEEE/CVF conference on computer vision and
  pattern recognition}, pages 12873--12883, 2021.

\bibitem{ArtifactUseFreq}
Joel Frank, Thorsten Eisenhofer, Lea Sch\"{o}nherr, Asja Fischer, Dorothea
  Kolossa, and Thorsten Holz.
\newblock Leveraging frequency analysis for deep fake image recognition.
\newblock In {\em Proceedings of the 37th International Conference on Machine
  Learning}, ICML'20. JMLR.org, 2020.

\bibitem{GAN}
Ian Goodfellow, Jean Pouget-Abadie, Mehdi Mirza, Bing Xu, David Warde-Farley,
  Sherjil Ozair, Aaron Courville, and Y.~Bengio.
\newblock Generative adversarial networks.
\newblock {\em Advances in Neural Information Processing Systems}, 3, 06 2014.

\bibitem{FakeOrJPEG}
Patrick Grommelt, Louis Weiss, Franz-Josef Pfreundt, and Janis Keuper.
\newblock Fake or jpeg? revealing common biases in generated image detection
  datasets.
\newblock {\em ArXiv}, abs/2403.17608, 2024.

\bibitem{VQDM}
Shuyang Gu, Dong Chen, Jianmin Bao, Fang Wen, Bo~Zhang, Dongdong Chen, Lu~Yuan,
  and Baining Guo.
\newblock Vector quantized diffusion model for text-to-image synthesis.
\newblock In {\em Proceedings of the IEEE/CVF Conference on Computer Vision and
  Pattern Recognition}, pages 10696--10706, 2022.

\bibitem{deepseek-r1}
Daya Guo, Dejian Yang, Haowei Zhang, Junxiao Song, Ruoyu Zhang, Runxin Xu,
  Qihao Zhu, Shirong Ma, Peiyi Wang, Xiao Bi, et~al.
\newblock Deepseek-r1: Incentivizing reasoning capability in llms via
  reinforcement learning.
\newblock {\em arXiv preprint arXiv:2501.12948}, 2025.

\bibitem{Infinity}
Jian Han, Jinlai Liu, Yi~Jiang, Bin Yan, Yuqi Zhang, Zehuan Yuan, Bingyue Peng,
  and Xiaobing Liu.
\newblock Infinity: Scaling bitwise autoregressive modeling for high-resolution
  image synthesis.
\newblock {\em arXiv preprint arXiv:2412.04431}, 2024.

\bibitem{DDPM}
Jonathan Ho, Ajay Jain, and P.~Abbeel.
\newblock Denoising diffusion probabilistic models.
\newblock {\em ArXiv}, abs/2006.11239, 2020.

\bibitem{Vision-R1}
Wenxuan Huang, Bohan Jia, Zijie Zhai, Shaosheng Cao, Zheyu Ye, Fei Zhao, Zhe
  Xu, Yao Hu, and Shaohui Lin.
\newblock Vision-r1: Incentivizing reasoning capability in multimodal large
  language models.
\newblock {\em arXiv preprint arXiv:2503.06749}, 2025.

\bibitem{openai-o1}
Aaron Jaech, Adam Kalai, Adam Lerer, Adam Richardson, Ahmed El-Kishky, Aiden
  Low, Alec Helyar, Aleksander Madry, Alex Beutel, Alex Carney, et~al.
\newblock Openai o1 system card.
\newblock {\em arXiv preprint arXiv:2412.16720}, 2024.

\bibitem{T2I-R1}
Dongzhi Jiang, Ziyu Guo, Renrui Zhang, Zhuofan Zong, Hao Li, Le~Zhuo, Shilin
  Yan, Pheng-Ann Heng, and Hongsheng Li.
\newblock T2i-r1: Reinforcing image generation with collaborative
  semantic-level and token-level cot.
\newblock {\em arXiv preprint arXiv:2505.00703}, 2025.

\bibitem{StyleGAN}
Tero Karras, Samuli Laine, and Timo Aila.
\newblock A style-based generator architecture for generative adversarial
  networks.
\newblock {\em 2019 IEEE/CVF Conference on Computer Vision and Pattern
  Recognition (CVPR)}, pages 4396--4405, 2018.

\bibitem{FLUX}
Black~Forest Labs.
\newblock Flux.
\newblock \url{https://github.com/black-forest-labs/flux}, 2024.

\bibitem{FakeBench}
Yixuan Li, Xuelin Liu, Xiaoyang Wang, Shiqi Wang, and Weisi Lin.
\newblock Fakebench: Uncover the achilles' heels of fake images with large
  multimodal models.
\newblock {\em ArXiv}, abs/2404.13306, 2024.

\bibitem{ArtifactTextureDefake}
Zhengzhe Liu, Xiaojuan Qi, and Philip~H.S. Torr.
\newblock Global texture enhancement for fake face detection in the wild.
\newblock In {\em 2020 IEEE/CVF Conference on Computer Vision and Pattern
  Recognition (CVPR)}, pages 8057--8066, 2020.

\bibitem{IntrinsicDimensionalities}
Peter Lorenz, Ricard Durall, and Janis Keuper.
\newblock Detecting images generated by deep diffusion models using their local
  intrinsic dimensionality.
\newblock {\em 2023 IEEE/CVF International Conference on Computer Vision
  Workshops (ICCVW)}, pages 448--459, 2023.

\bibitem{Midjourney}
Midjourney.
\newblock Midjourney, 2023.

\bibitem{GLIDE}
Alex Nichol, Prafulla Dhariwal, Aditya Ramesh, Pranav Shyam, Pamela Mishkin,
  Bob McGrew, Ilya Sutskever, and Mark Chen.
\newblock Glide: Towards photorealistic image generation and editing with
  text-guided diffusion models.
\newblock {\em arXiv preprint arXiv:2112.10741}, 2021.

\bibitem{NPR}
Utkarsh Ojha, Yuheng Li, and Yong~Jae Lee.
\newblock Towards universal fake image detectors that generalize across
  generative models.
\newblock {\em 2023 IEEE/CVF Conference on Computer Vision and Pattern
  Recognition (CVPR)}, pages 24480--24489, 2023.

\bibitem{DALLE3}
OpenAI.
\newblock Dall·e 3 system card, 2023.

\bibitem{OpenAI4oGen}
OpenAI.
\newblock Introducing 4o image generation, Mar 2025.

\bibitem{SFT}
Long Ouyang, Jeffrey Wu, Xu~Jiang, Diogo Almeida, Carroll Wainwright, Pamela
  Mishkin, Chong Zhang, Sandhini Agarwal, Katarina Slama, Alex Ray, et~al.
\newblock Training language models to follow instructions with human feedback.
\newblock {\em Advances in neural information processing systems},
  35:27730--27744, 2022.

\bibitem{MedVLM-R1}
Jiazhen Pan, Che Liu, Junde Wu, Fenglin Liu, Jiayuan Zhu, Hongwei~Bran Li, Chen
  Chen, Cheng Ouyang, and Daniel Rueckert.
\newblock Medvlm-r1: Incentivizing medical reasoning capability of
  vision-language models (vlms) via reinforcement learning.
\newblock {\em arXiv preprint arXiv:2502.19634}, 2025.

\bibitem{ComFor}
Jeongsoo Park and Andrew Owens.
\newblock Community forensics: Using thousands of generators to train fake
  image detectors, 2024.

\bibitem{DiT}
William Peebles and Saining Xie.
\newblock Scalable diffusion models with transformers.
\newblock In {\em Proceedings of the IEEE/CVF international conference on
  computer vision}, pages 4195--4205, 2023.

\bibitem{LMM-R1}
Yingzhe Peng, Gongrui Zhang, Miaosen Zhang, Zhiyuan You, Jie Liu, Qipeng Zhu,
  Kai Yang, Xingzhong Xu, Xin Geng, and Xu~Yang.
\newblock Lmm-r1: Empowering 3b lmms with strong reasoning abilities through
  two-stage rule-based rl.
\newblock {\em arXiv preprint arXiv:2503.07536}, 2025.

\bibitem{Qwen25VL}
{Qwen Team}.
\newblock Qwen2.5-vl, January 2025.

\bibitem{DALLE2}
Aditya Ramesh, Prafulla Dhariwal, Alex Nichol, Casey Chu, and Mark Chen.
\newblock Hierarchical text-conditional image generation with clip latents.
\newblock {\em arXiv preprint arXiv:2204.06125}, 1(2):3, 2022.

\bibitem{Dataset4oPreference}
Rapidata.
\newblock Rapidata openai 4o preference, Mar 2025.

\bibitem{AEROBLADE}
Jonas Ricker, Denis Lukovnikov, and Asja Fischer.
\newblock Aeroblade: Training-free detection of latent diffusion images using
  autoencoder reconstruction error.
\newblock {\em 2024 IEEE/CVF Conference on Computer Vision and Pattern
  Recognition (CVPR)}, pages 9130--9140, 2024.

\bibitem{SD21}
Robin Rombach, Andreas Blattmann, Dominik Lorenz, Patrick Esser, and Bj\"orn
  Ommer.
\newblock High-resolution image synthesis with latent diffusion models.
\newblock In {\em Proceedings of the IEEE/CVF Conference on Computer Vision and
  Pattern Recognition (CVPR)}, pages 10684--10695, June 2022.

\bibitem{SD14}
Robin Rombach, Andreas Blattmann, Dominik Lorenz, Patrick Esser, and Bj\"orn
  Ommer.
\newblock High-resolution image synthesis with latent diffusion models.
\newblock In {\em Proceedings of the IEEE/CVF Conference on Computer Vision and
  Pattern Recognition (CVPR)}, pages 10684--10695, June 2022.

\bibitem{FaceForensicsPP}
Andreas R\"ossler, Davide Cozzolino, Luisa Verdoliva, Christian Riess, Justus
  Thies, and Matthias Nie{\ss}ner.
\newblock Face{F}orensics++: Learning to detect manipulated facial images.
\newblock In {\em International Conference on Computer Vision (ICCV)}, 2019.

\bibitem{Imagen}
Chitwan Saharia, William Chan, Saurabh Saxena, Lala Li, Jay Whang, Emily~L
  Denton, Kamyar Ghasemipour, Raphael Gontijo~Lopes, Burcu Karagol~Ayan, Tim
  Salimans, et~al.
\newblock Photorealistic text-to-image diffusion models with deep language
  understanding.
\newblock {\em Advances in Neural Information Processing Systems},
  35:36479--36494, 2022.

\bibitem{XplainedGradCAM}
Ramprasaath~R. Selvaraju, Michael Cogswell, Abhishek Das, Ramakrishna Vedantam,
  Devi Parikh, and Dhruv Batra.
\newblock Grad-cam: Visual explanations from deep networks via gradient-based
  localization.
\newblock In {\em 2017 IEEE International Conference on Computer Vision
  (ICCV)}, pages 618--626, 2017.

\bibitem{DeepSeekMath}
Zhihong Shao, Peiyi Wang, Qihao Zhu, Runxin Xu, Jun-Mei Song, Mingchuan Zhang,
  Y.~K. Li, Yu~Wu, and Daya Guo.
\newblock Deepseekmath: Pushing the limits of mathematical reasoning in open
  language models.
\newblock {\em ArXiv}, abs/2402.03300, 2024.

\bibitem{VLM-R1}
Haozhan Shen, Peng Liu, Jingcheng Li, Chunxin Fang, Yibo Ma, Jiajia Liao,
  Qiaoli Shen, Zilun Zhang, Kangjia Zhao, Qianqian Zhang, Ruochen Xu, and
  Tiancheng Zhao.
\newblock Vlm-r1: A stable and generalizable r1-style large vision-language
  model.
\newblock {\em arXiv preprint arXiv:2504.07615}, 2025.

\bibitem{XplainedCNN}
Karen Simonyan, Andrea Vedaldi, and Andrew Zisserman.
\newblock Deep inside convolutional networks: Visualising image classification
  models and saliency maps.
\newblock {\em CoRR}, abs/1312.6034, 2013.

\bibitem{DDIM}
Jiaming Song, Chenlin Meng, and Stefano Ermon.
\newblock Denoising diffusion implicit models.
\newblock {\em ArXiv}, abs/2010.02502, 2020.

\bibitem{LlamaGen}
Peize Sun, Yi~Jiang, Shoufa Chen, Shilong Zhang, Bingyue Peng, Ping Luo, and
  Zehuan Yuan.
\newblock Autoregressive model beats diffusion: Llama for scalable image
  generation.
\newblock {\em arXiv preprint arXiv:2406.06525}, 2024.

\bibitem{XplainedAxiomaticAttr}
Mukund Sundararajan, Ankur Taly, and Qiqi Yan.
\newblock Axiomatic attribution for deep networks.
\newblock In {\em Proceedings of the 34th International Conference on Machine
  Learning - Volume 70}, ICML'17, page 3319–3328. JMLR.org, 2017.

\bibitem{EfficientNet}
Mingxing Tan and Quoc~V. Le.
\newblock Efficientnet: Rethinking model scaling for convolutional neural
  networks.
\newblock {\em ArXiv}, abs/1905.11946, 2019.

\bibitem{GALIP}
Ming Tao, Bing-Kun Bao, Hao Tang, and Changsheng Xu.
\newblock Galip: Generative adversarial clips for text-to-image synthesis.
\newblock In {\em Proceedings of the IEEE/CVF Conference on Computer Vision and
  Pattern Recognition}, pages 14214--14223, 2023.

\bibitem{VAR}
Keyu Tian, Yi~Jiang, Zehuan Yuan, Bingyue Peng, and Liwei Wang.
\newblock Visual autoregressive modeling: Scalable image generation via
  next-scale prediction.
\newblock {\em Advances in neural information processing systems},
  37:84839--84865, 2024.

\bibitem{VQVAE}
Aaron Van Den~Oord, Oriol Vinyals, et~al.
\newblock Neural discrete representation learning.
\newblock {\em Advances in neural information processing systems}, 30, 2017.

\bibitem{VL-Rethinker}
Haozhe Wang, Chao Qu, Zuming Huang, Wei Chu, Fangzhen Lin, and Wenhu Chen.
\newblock Vl-rethinker: Incentivizing self-reflection of vision-language models
  with reinforcement learning.
\newblock {\em arXiv preprint arXiv:2504.08837}, 2025.

\bibitem{ObjectFormer}
Junke Wang, Zuxuan Wu, Jingjing Chen, Xintong Han, Abhinav Shrivastava, Ser-Nam
  Lim, and Yu-Gang Jiang.
\newblock Objectformer for image manipulation detection and localization.
\newblock In {\em Proceedings of the IEEE/CVF Conference on Computer Vision and
  Pattern Recognition}, 2022.

\bibitem{Qwen2VL}
Peng Wang, Shuai Bai, Sinan Tan, Shijie Wang, Zhihao Fan, Jinze Bai, Keqin
  Chen, Xuejing Liu, Jialin Wang, Wenbin Ge, Yang Fan, Kai Dang, Mengfei Du,
  Xuancheng Ren, Rui Men, Dayiheng Liu, Chang Zhou, Jingren Zhou, and Junyang
  Lin.
\newblock Qwen2-vl: Enhancing vision-language model's perception of the world
  at any resolution.
\newblock {\em arXiv preprint arXiv:2409.12191}, 2024.

\bibitem{CNNSpot}
Sheng-Yu Wang, Oliver Wang, Richard Zhang, Andrew Owens, and Alexei~A Efros.
\newblock Cnn-generated images are surprisingly easy to spot... for now.
\newblock In {\em Proceedings of the IEEE/CVF conference on computer vision and
  pattern recognition}, pages 8695--8704, 2020.

\bibitem{DIRE}
Zhendong Wang, Jianmin Bao, Wengang Zhou, Weilun Wang, Hezhen Hu, Hong Chen,
  and Houqiang Li.
\newblock Dire for diffusion-generated image detection.
\newblock In {\em Proceedings of the IEEE/CVF International Conference on
  Computer Vision}, pages 22445--22455, 2023.

\bibitem{Fast-Slow-Thinking}
Wenyi Xiao, Leilei Gan, Weilong Dai, Wanggui He, Ziwei Huang, Haoyuan Li,
  Fangxun Shu, Zhelun Yu, Peng Zhang, Hao Jiang, et~al.
\newblock Fast-slow thinking for large vision-language model reasoning.
\newblock {\em arXiv preprint arXiv:2504.18458}, 2025.

\bibitem{SegFormer}
Enze Xie, Wenhai Wang, Zhiding Yu, Anima Anandkumar, Jose~M Alvarez, and Ping
  Luo.
\newblock Segformer: Simple and efficient design for semantic segmentation with
  transformers.
\newblock In {\em Neural Information Processing Systems (NeurIPS)}, 2021.

\bibitem{R1-OneVision}
Yi~Yang, Xiaoxuan He, Hongkun Pan, Xiyan Jiang, Yan Deng, Xingtao Yang, Haoyu
  Lu, Dacheng Yin, Fengyun Rao, Minfeng Zhu, et~al.
\newblock R1-onevision: Advancing generalized multimodal reasoning through
  cross-modal formalization.
\newblock {\em arXiv preprint arXiv:2503.10615}, 2025.

\bibitem{ABench}
Zicheng Zhang, Haoning Wu, Chunyi Li, Yingjie Zhou, Wei Sun, Xiongkuo Min,
  Zijian Chen, Xiaohong Liu, Weisi Lin, and Guangtao Zhai.
\newblock A-bench: Are lmms masters at evaluating ai-generated images?
\newblock {\em arXiv preprint arXiv:2406.03070}, 2024.

\bibitem{ms-swift}
Yuze Zhao, Jintao Huang, Jinghan Hu, Xingjun Wang, Yunlin Mao, Daoze Zhang,
  Zeyinzi Jiang, Zhikai Wu, Baole Ai, Ang Wang, et~al.
\newblock Swift: a scalable lightweight infrastructure for fine-tuning.
\newblock In {\em Proceedings of the AAAI Conference on Artificial
  Intelligence}, volume~39, pages 29733--29735, 2025.

\bibitem{VisualThinker-R1-Zero}
Hengguang Zhou, Xirui Li, Ruochen Wang, Minhao Cheng, Tianyi Zhou, and Cho-Jui
  Hsieh.
\newblock R1-zero's" aha moment" in visual reasoning on a 2b non-sft model.
\newblock {\em arXiv preprint arXiv:2503.05132}, 2025.

\end{thebibliography}
